\documentclass[conference]{IEEEtran}
\IEEEoverridecommandlockouts
\usepackage{amsmath,amssymb,amsfonts}
\usepackage{algorithmic}
\usepackage{hyperref}
\usepackage{booktabs}  
\usepackage{comment}
\usepackage{multirow}
\usepackage{graphicx}
\usepackage{textcomp}
\usepackage{xcolor}
\usepackage{cite}
\def\BibTeX{{\rm B\kern-.05em{\sc i\kern-.025em b}\kern-.08em
    T\kern-.1667em\lower.7ex\hbox{E}\kern-.125emX}}
\begin{document}

\title{Cross-Model Transferability of Adversarial Patches in Real-time Segmentation for Autonomous Driving\\
\thanks{This work was supported by the US Department of Transportation (USDOT) Tier-1 University Transportation Center (UTC) Transportation Cybersecurity Center for Advanced Research and Education (CYBER-CARE) (Grant No. 69A3552348332)}
}

\author{\IEEEauthorblockN{Prashant Shekhar}
\IEEEauthorblockA{\textit{Department of Mathematics} \\
\textit{Embry-Riddle Aeronautical University}\\
Daytona Beach, FL, USA \\
shekharp@erau.edu}
\and
\IEEEauthorblockN{Bidur Devkota}
\IEEEauthorblockA{\textit{Department of Mathematics} \\
\textit{Embry-Riddle Aeronautical University}\\
Daytona Beach, FL, USA \\
devkotab@my.erau.edu}
\and
\IEEEauthorblockN{Dumindu Samaraweera}
\IEEEauthorblockA{\textit{Department of Mathematics} \\
\textit{Embry-Riddle Aeronautical University}\\
Daytona Beach, FL, USA \\
samarawg@erau.edu}
\and
\IEEEauthorblockN{Laxima Niure Kandel}
\IEEEauthorblockA{\textit{Department of Electrical Engineering and Computer Science
} \\
\textit{Embry-Riddle Aeronautical University}\\
Daytona Beach, FL, USA \\
niurekal@erau.edu}
\and
\IEEEauthorblockN{Manoj Babu}
\IEEEauthorblockA{{WMG} \\
\textit{University of Warwick}\\
Warwick, UK \\
m.babu@warwick.ac.uk}
}

\maketitle

\begin{abstract}
Adversarial attacks pose a significant threat to deep learning models, particularly in safety-critical applications like healthcare and autonomous driving. Recently, patch based attacks have demonstrated effectiveness in real-time inference scenarios owing to their `drag and drop' nature. Following this idea for Semantic Segmentation (SS), here we propose a novel Expectation Over Transformation (EOT) based adversarial patch attack that is more realistic for autonomous vehicles. To effectively train this attack we also propose a `simplified' loss function that is easy to analyze and implement. Using this attack as our basis, we investigate whether adversarial patches once optimized on a specific SS model, can fool other models/architectures. We conduct a comprehensive cross-model transferability analysis of adversarial patches trained on state-of-the-art (SOTA) Convolutional Neural Network (CNN) models such PIDNet-S, PIDNet-M and PIDNet-L, among others. Additionally, we also include the Segformer model to study transferability to Vision Transformers (ViTs). All of our analysis is conducted on the widely used Cityscapes dataset. Our study reveals key insights into how model architectures (CNN vs CNN or CNN vs. Transformer-based) influence attack susceptibility. In particular, we conclude that although the transferability (effectiveness) of attacks on unseen images of any dimension is really high, the attacks trained against one particular model are minimally effective on other models. And this was found to be true for both ViT and CNN based models. Additionally our results also indicate that for CNN-based models, the repercussions of patch attacks are local, unlike ViTs.  Per-class analysis reveals that \textit{simple-classes} like `sky' suffer less misclassification than others. The code for the project is available at: \href{https://github.com/p-shekhar/adversarial-patch-transferability}{https://github.com/p-shekhar/adversarial-patch-transferability}
\end{abstract}

\begin{IEEEkeywords}
Adversarial patches, Robustness, Semantic segmentation, Intelligent transport systems, Autonomous driving
\end{IEEEkeywords}

\section{Introduction}
Deep learning has transformed computer vision, enabling significant advancements in tasks such as image classification, object detection, and semantic segmentation \cite{zhao2024review,chen2024survey}. Among these, real-time semantic segmentation plays a crucial role in various applications, including autonomous driving \cite{xu2023pidnet}, robotics \cite{yan2020roboseg}, medical imaging \cite{awasthi2024real}, and surveillance \cite{pierard2023mixture}. The ability to accurately label each pixel in an image allows autonomous systems to understand their surroundings, make informed decisions, and operate safely in dynamic environments. However, as these systems become increasingly reliant on deep learning models, their vulnerability to adversarial attacks has emerged as a critical security concern \cite{akhtar2018threat,badjie2024adversarial,costa2024deep}.
\begin{figure}
    \centering
    \includegraphics[width=1\linewidth]{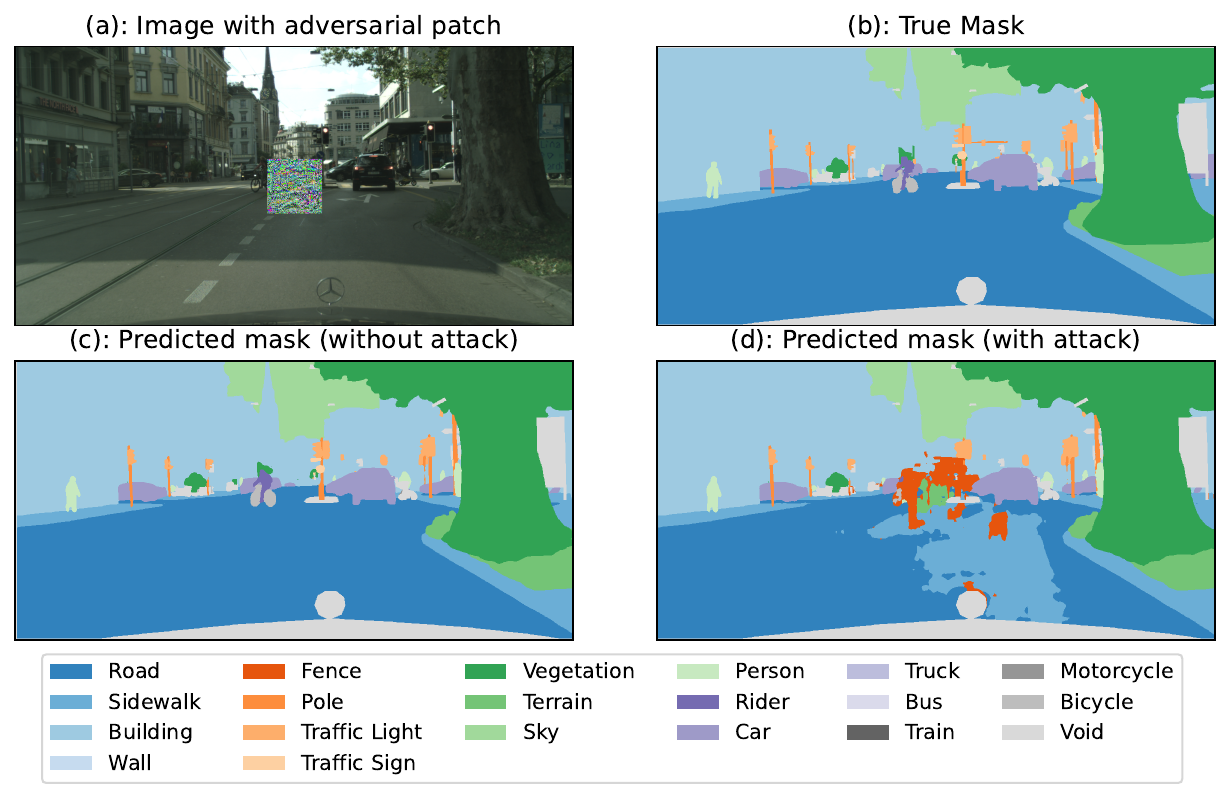}
    \caption{Demonstrating the impact of adversarial patch attacks. (a) shows an optimized patch attack for Cityscapes dataset with respect to pretrained PIDNet-L model. (b) shows the true segmentation mask and (c) shows the predicted segmentation mask in absence of any attack. (d) shows the predicted mask when the attacked image (shown in (a)) is passed through the same pretrained PIDNet-L model}
    \label{fig:enter-label}
\end{figure}
Adversarial attacks involve carefully crafted perturbations designed to deceive neural networks into making incorrect predictions. While traditional pixel-wise perturbation-based attacks have been extensively studied \cite{fang2024one,li2024adversarial}, they often require noise spread across the entire image, making them difficult to apply in real-world scenarios. In this regard, attacks such as Fast-Gradient Sign Method (FGSM) \cite{goodfellow2014explaining} add imperceptible noise to the entire image learnt using a particular model, thereby forcing the model to make prediction errors on that image. However in autonomous driving scenarios, such carefully crafted, image-specific noise filters do not generalize to other images captured at different angles and lightening, rendering them in-effective. In contrast, patch-based adversarial attacks introduce highly visible yet localized perturbations, which can be physically printed and placed within a scene. These adversarial patches have been shown to effectively fool classification and detection models, raising concerns about their impact on semantic segmentation models \cite{rossolini2023real}, particularly in safety-critical applications. As a demonstration, here Fig. \ref{fig:enter-label}(a) presents the application of an adversarial patch trained on images from the Cityscapes dataset \cite{cordts2016cityscapes} using the PIDNet-L \cite{xu2023pidnet} SS model for evaluation. More details about the patch training are provided in the following sections. To notice the effectiveness of the patch attack, readers are advised to look at the segmentation mask predicted from the attacked image (shown in Fig. \ref{fig:enter-label}(d)). Here, based on the legends we can see the patch is forcing the pretrained model (that inherently predict well as shown in Fig. \ref{fig:enter-label}(c)) to predict incorrectly on pixels that are not even covered by the patch. In-fact such attacks are particularly effective since the model is forced to predict classes like `fence' or `sidewalk' in the middle of the road leading to events such as emergency braking or abrupt turning, causing accidents.

\subsection{Adversarial Patch Attacks in Segmentation Models}

Typically, adversarial patch attacks in classification models focus on changing the output label of an image, hence the attack is responsible for just changing a scalar prediction \cite{brown2017adversarial}. However for segmentation models where the requirement is spatially consistent predictions across an entire scene, producing effective and efficient adversarial attacks is difficult. The reasoning behind this is the requirement for simultaneously fooling the model on multiple pixels across the scene. This makes adversarial attacks on segmentation models more challenging but also more impactful in real-world applications \cite{rony2023proximal}. For instance, an attack could manipulate an autonomous vehicle’s perception system by making pedestrians or road signs disappear, or as shown in Fig \ref{fig:enter-label} can lead to appearance of a fence on the road, leading to catastrophic consequences. Adversarial patches designed for segmentation models can disrupt the spatial coherence of segmentation masks, causing severe misclassification of critical regions \cite{arnab2018robustness}.

One of the key open questions in adversarial robustness is the transferability of adversarial patches across models. Transferability refers to the ability of an adversarial patch trained on one model to successfully deceive other models without additional optimization. If adversarial patches are highly transferable, an attacker could design a single patch that simultaneously degrades the performance of multiple segmentation architectures, reducing the effectiveness of model diversity as a defense strategy. Understanding how different architectures react to adversarial patches is crucial for designing more resilient segmentation models and robust defense mechanisms.

\subsection{Real-Time Semantic Segmentation Models and Their Vulnerabilities}
Real-time semantic segmentation models are designed to balance accuracy and computational efficiency, enabling their deployment in resource-constrained environments such as embedded vision systems and autonomous vehicles. Modern SS architectures can be broadly categorized into two types:

\begin{itemize}
    \item \textit{CNN-based models}: These include architectures such as PIDNet \cite{xu2023pidnet}, BiSeNet \cite{yu2018bisenet,yu2021bisenet}, and ICNet \cite{zhao2018icnet}, which leverage convolutional layers for hierarchical feature extraction. CNN-based models are highly optimized for efficiency, making them well-suited for real-time applications.
    \item \textit{Transformer-based models}: Recent advances in Vision Transformers (ViTs) \cite{han2022survey} have introduced architectures such as SegFormer \cite{xie2021segformer}, which utilize self-attention mechanisms to capture long-range dependencies. Transformers have demonstrated superior performance in many computer vision tasks, but their robustness to adversarial attacks remains an area of active research.
\end{itemize}

Despite their efficiency and high performance, real-time segmentation models remain susceptible to adversarial perturbations. CNN-based models, while optimized for speed, often rely on spatially localized feature extraction, making them potentially vulnerable to small, well-placed adversarial patches. On the other hand, Transformer-based models process global context and may exhibit different adversarial characteristics, either being more robust due to self-attention mechanisms or more vulnerable to specific structured attacks.

This study investigates the cross-model transferability of adversarial patches across a diverse set of real-time segmentation architectures. In this research we aim to answer the following key research questions (i): Can an adversarial patch trained on one segmentation model successfully fool other models ?
(ii): Are CNN-based models and Transformer-based models equally susceptible to adversarial patches ? (iii): Which semantic classes (e.g., roads, pedestrians, vehicles) are most affected by black-box adversarial perturbations ? (iv): How does model architecture influence adversarial robustness and transferability?

\subsection{Contributions of This Work}
To address these questions, we conduct a systematic evaluation of adversarial patch attacks on multiple semantic segmentation models, covering both CNN-based architectures (PIDNet-S, PIDNet-M, PIDNet-L, ICNet, BiSeNetV1, BiSeNetV2) and a ViT-based architecture (SegFormer). Here all CNN based architectures are strictly real-time SS Models. However Segformer on the other does not belong to this class. Still we have included Segformer in this research to evaluate the CNN vs ViT attack scenarios. Overall, our contributions are as follows:

\begin{enumerate}
\item \textit{Novel Expectation Over Transformation (EOT) based attack formulation}: We propose a variant of the original EOT formulation \cite{athalye2018synthesizing} for learning blackbox adversarial patch attacks, thus making it more realistic for a Real-time SS model in autonomous driving scenarios.
\item \textit{Adaptive adversarial training loss}: For learning the patch itself, we proposed an adaptive loss function that simplifies the one introduced in \cite{rossolini2023real}. In particular, we reduce the number of hyper-parameters in the loss metric to make the attack more robust.
    \item \textit{Comprehensive Transferability Study}: 
We analyze how well adversarial patches transfer across different segmentation models, identifying key architectural weaknesses. Additionally, by comparing CNN-based and ViT-based models, we provide insights into their relative robustness against patch-based attacks. To add to this, we also evaluate the per-class performance degradation to determine which object categories (e.g., pedestrians, vehicles, buildings) are most affected by adversarial patches. \textit{To the best of our knowledge, this is the first research to compare the performance of ViTs and CNN based SS models against patch based adversarial attacks at this level of detail in autonomous driving domain}.
\end{enumerate}

Our findings (sections \ref{decay} and \ref{transferability}) highlight the security risks posed by adversarial patches in real-time segmentation models and emphasize the need for robust defenses against such attacks. The insights gained from this study will help guide future research in adversarial robustness, defense mechanisms, and secure deployment of deep learning models in real-world applications.

\section{Attack Model}
As mentioned in the previous section, here we build on the original formulation of EOT based attacks proposed in \cite{athalye2018synthesizing}. For classification problems, such attacks can be formulated as:

\begin{equation}
    x' = \arg\max_{x'} \mathbb{E}_{t \sim T} \left[ \log P(y_t | t(x')) \right]
    \label{old}
\end{equation}
subject to:
\[
\mathbb{E}_{t \sim T} \left[d(t(x'),t(x)) \right] < \epsilon, \quad x, x' \in [0,1]^d
\]

where:
\begin{itemize}
    \item $x$ is the image under consideration having $d$ pixels
    \item \( x' = x  +\delta \) is the adversarial image produced after optimizing $\delta$.
    \item \( T \) is a distribution over transformations of interest.
    \item \( t(x') = t(x + \delta) \) is a transformation function (\( t \sim T \)) applied to the adversarial example.
    \item \( P(y_t | t(x')) \) denotes the probability of the target class \( y_t \) given the transformed adversarial input.
    \item $d(\cdot,\cdot)$ is a distance metric quantifying the discrepancy between original and attacked transformed image.
\end{itemize}

The optimization process in (\ref{old}) ensures that the adversarial perturbation is robust to a set of transformations (for example $t \sim T$), making it effective in real-world conditions. In other words, when the image $x$ is optimally attacked to form $x'$, then irrespective of the transformations applied over it (from a fixed family of transformations), on average the attacked model will predict the wrong class $y_t$ instead of the true class label $y$ for $x$. In this research however, we move away from the targeted setting and focus on untargeted attacks in SS, where the objective is to force the model to predict incorrectly without having a target class of interest. To implement this, we minimize the probability of the true class associated with a pixel of the image. But we do this under the EOT framework. Assuming $(x,y) \sim P_{xy}(X,Y)$, where x is an image and y is its corresponding segmentation mask. Additionally, we assume a joint data distribution $P_{xy}$. Using these notations, we solve the following optimization problem:

\begin{equation}
    \delta^{*} = \arg\min_{\delta} \mathbb{E}_{(x,y) \sim P_{xy}} \mathbb{E}_{t \sim T} \left[ \log P(y | t(x')) \right]
    \label{new}
\end{equation}
subject to:
\[
\mathbb{E}_{t \sim T} \left[d(t(x'),t(x)) \right] < \epsilon, \quad x \in [0,1]^d
\]
\[
t(x') = t(x)\circ I + P\delta
\]
It should be noted that (\ref{new}) is a minimization problem using the true label $y$, as opposed to the maximization problem involving target classification label $y_t$ in (\ref{old}). One additional key difference is that in formulation (\ref{new}) we assume $x' = x \circ I + P\delta$, where $\circ$ is an elementwise multiplication operator and $x \circ I$ zeros out the pixel values in the location where an adversarial patch is to be added. Also, $\delta$ is the patch of adversarial pixels, and the operator $P$ pads 0's to this patch to match the dimensions of $x$. Furthermore, during model training we incorporate transformations as: $t(x') = t(x)\circ I + P\delta$. Hence computationally, we apply transformation on the clean image and then paste the patch on it to create a transformed adversarial example. This is different than $t(x') = t(x + \delta)$ in (\ref{old}) where transformation was applied on an already attacked image. $t(x') = t(x)\circ I + P\delta$ is more suitable for training a generalizable patch attack since in real driving scenarios the images captured by the car camera would be at different angles, distorted, and in different lighting etc. But the attack should be independent of those transformations and should work in its standard form.

\begin{figure*}
    \centering
    \includegraphics[width=0.80\linewidth]{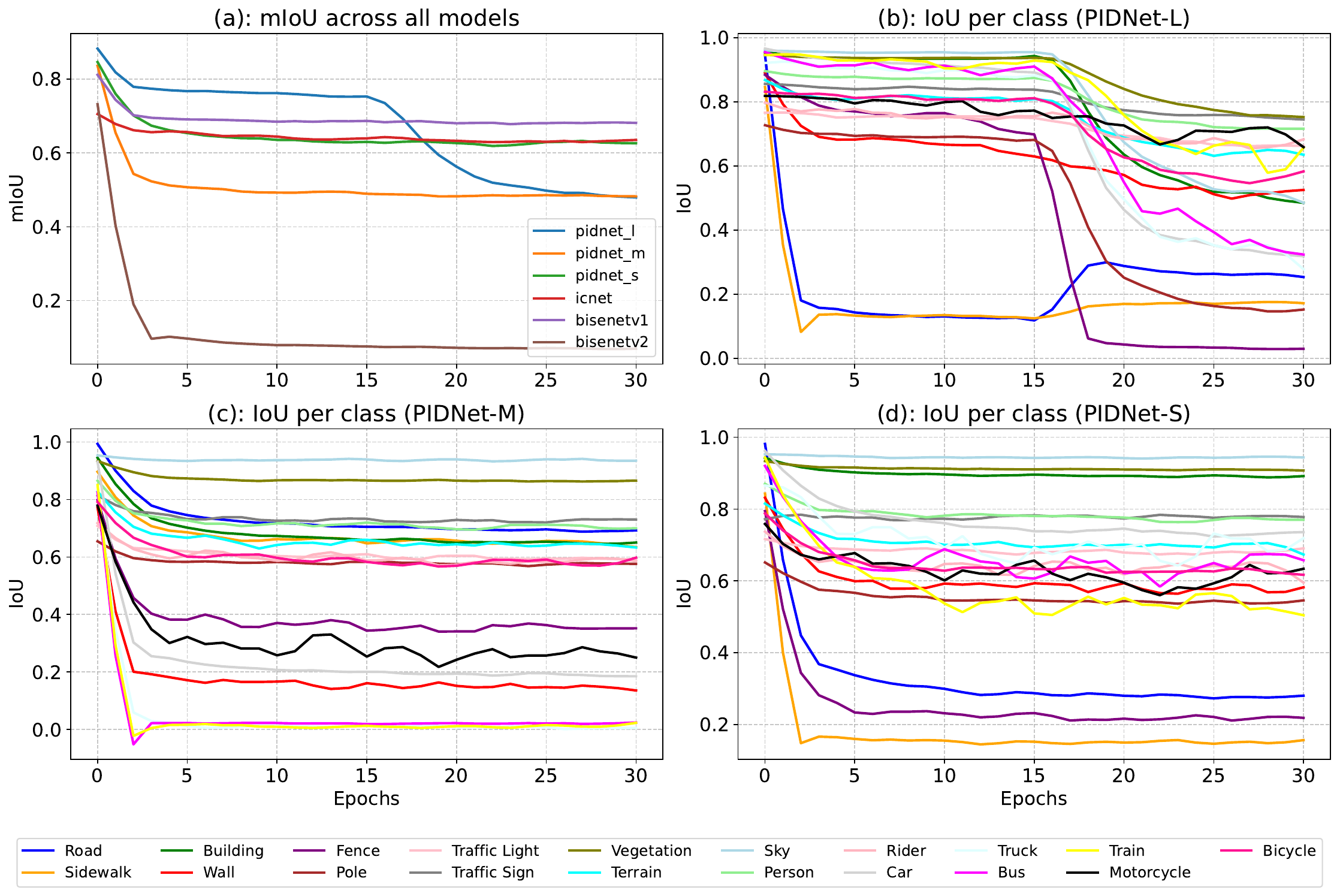}
    \caption{(a): MIoU decay of real-time SS models as model-specific adversarial patches are trained on the Cityscapes-train dataset. (b,c,d): Per-class Intersection over Union (IoU) decay during patch training on the Cityscapes-train dataset with respect to PIDNet family of models. Class legends are shown at the bottom}
    \label{fig:enter-label2}
\end{figure*}

To understand the adversarial patch learning process consider a pretrained SS model $m$. For an image pixel $x_i$ if the model correctly predicts the class, then we will have $y_i = m(x_i)$. Additionally let $\Omega_x$ be the set of correctly classified pixels whereas $\Omega^c_x$ is the set of pixels incorrectly classified. Hence the set of all pixels of an image $x$ would be denoted by the set: $\Delta_x = \Omega_x \cup \Omega^c_x \cup \delta$. Assuming these notations, for training an adversarial patch (using a batch of $n$ images) we exclusively focus on the cross-entropy loss of all pixels that are correctly classified, i.e.
\begin{equation}
    L = (1/n)\sum_i L_{x_i},\text{ where } L_x = -(1/|\Omega_x|)\sum_{i \in \Omega_x} y_i log(m(x_i))
    \label{loss}
\end{equation}
This is a simplification of the previous works such as \cite{rossolini2023real} that considered both the losses of correctly and incorrectly classified pixels and weighted them according to a hyper-parameter $\gamma$. However to keep the loss function simple, in this research we only use loss of correctly classified pixels ($\Omega_x$) to drive the optimization of patch $\delta$. During optimization we expect as the number of correctly classified pixels reduce, the total loss $L$ wouldn't go up considerably as (\ref{loss}) is the mean quantity over correctly classified pixels. However, still the Mean Intersection over Union (MIoU) on pixels in the set $\Omega \cup \Omega^c$ would reduce. This is achieved by the following gradient ascent step over the pixels in the patch ($\delta$):
\begin{equation}
    \delta \leftarrow \delta  + \varepsilon \cdot \operatorname{sign}(\nabla_{\delta}L)
    \label{SGA}
\end{equation}

Motivated from the FGSM algorithm \cite{goodfellow2014explaining}, here we use the sign function to ensure uniformity in gradient updates. To keep the attack realistic, we clip the $\delta$ pixels that go out of the range [0,1]. Additionally, it should be noted that algorithm (\ref{SGA}) coupled with pixel clipping ensures $t(x') = t(x)\circ I + P\delta$ satisfies the requirement $\mathbb{E}_{t \sim T} \left[d(t(x'),t(x)) \right] < \epsilon$.

\begin{table*}[h]
    \centering
    \renewcommand{\arraystretch}{1.2}  
    \setlength{\tabcolsep}{6pt}  
    \scalebox{0.9}{  
    \begin{tabular}{lccccccc}
        \toprule
        \multirow{2}{*}{\textbf{Attack Mode}} & \multicolumn{7}{c}{\textbf{Model-wise MIoU on attacked Cityscapes-val images}} \\  
        \cmidrule(lr){2-8} 
        & \textbf{PIDNet-L} & \textbf{PIDNet-M} & \textbf{PIDNet-S} & \textbf{ICNet} & \textbf{BiSeNetV1} & \textbf{BiSeNetV2} & \textbf{Segformer} \\
        \midrule
        \textbf{Random Patch}   & 0.8996  & 0.8619  & 0.8652  & 0.6555  & 0.7110  & 0.6844  & 0.8118 \\
        \midrule
        \textbf{PIDNet-L Patch} & 0.7914  & 0.8613  & 0.8633  & 0.6555  & 0.7118  & 0.6849  & 0.8101 \\
        \textbf{PIDNet-M Patch} & 0.8982  & 0.7316  & 0.8629  & 0.6548  & 0.7120  & 0.6846  & 0.8105 \\
        \textbf{PIDNet-S Patch} & 0.8984  & 0.8615  & 0.7791  & 0.6546  & 0.7120  & 0.6848  & 0.8105 \\
        \textbf{ICNet Patch}    & 0.8993  & 0.8616  & 0.8648  & 0.6204  & 0.7115  & 0.6849  & 0.8109 \\
        \textbf{BiSeNet V1 Patch}     & 0.8994  & 0.8613  & 0.8642  & 0.6553  & 0.6505  & 0.6844  & 0.8097 \\
        \textbf{BiSeNet V2 Patch}     & 0.8992  & 0.8616  & 0.8648  & 0.6553  & 0.7117  & 0.2991  & 0.8111 \\
        \textbf{Segformer Patch}      & 0.8997  & 0.8624  & 0.8652  & 0.6538  & 0.7135  & 0.6858  & 0.4491 \\
        \bottomrule
    \end{tabular}
    }
    \caption{Effectiveness comparison for different adversarial patch attacks evaluated on Cityscapes-val dataset.}
    \label{tab:cityscape_val}
\end{table*}

\begin{figure*}
    \centering
    \includegraphics[width=1\linewidth]{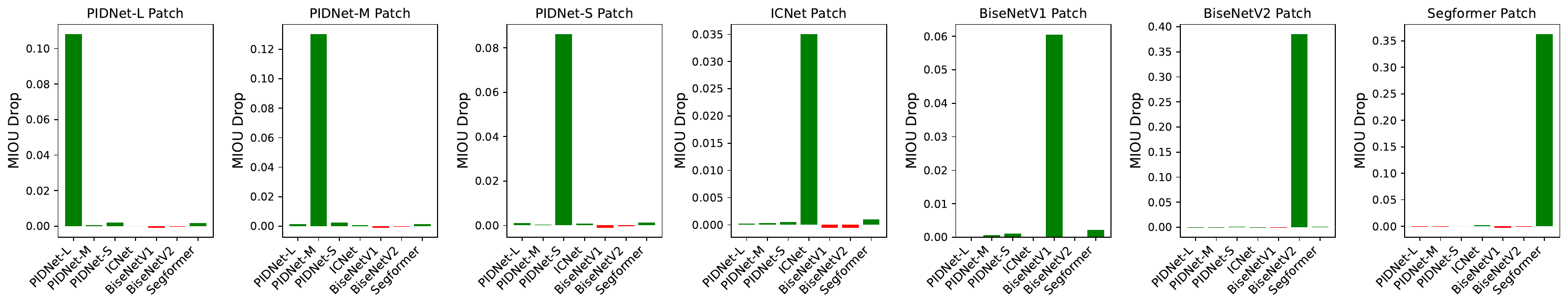}
    \caption{Analysis of performance drop with Intra-Model (trained-on and attack the same model) and Inter-Model (trained-on and attack different models) patch based attack discussed in this paper. Here individual subplots demonstrate a model-specific patch attack evaluated on all 7 models under consideration. These patch attacks were trained using Cityscapes-train dataset and evaluated using Cityscapes-val dataset.} 
    \label{fig:enter-label4}
\end{figure*}

\section{Experimental Results}
\label{Results}
In this section we present the results from our proposed approach, analyzing the robustness and transferability of the proposed patch attack. 
\subsection{Patch training setup}
We begin by obtaining the SS models of interest (PIDNet-S,-M,-L, ICNet, BiSeNetV1, BiSeNetV2 and Segformer) from \cite{PIDNet,ICNet,BiseNetV1V2,DBLP:journals/corr/abs-2105-15203}. Among these models PIDNet-S, -M, -L, BiSeNetV1, V2 and Segformer are already pre-trained on Cityscapes dataset. For ICNet, pretraining on Cityscapes is performed using the training scripts from PIDNet repository \cite{PIDNet}. In order to keep things simple we assume all patches are to be placed at the center of the images. During patch training we work with Cityscapes-train dataset that has 2,975 images with $1024\times 2048$ resolution. Regarding transformations $t\sim T$, we perform scaling of the images in the range [0.5,2], and then crop it to $1024\times 1024$. Additionally, we also flip the image horizontally to simulate different views. During training, after each image (in a batch) goes through these transformations we implement $t(x') = t(x)\circ I + P\delta$ by pasting the patch at the center. Using the output of SS model, we compute the loss using (\ref{loss}) and use the gradient ascent step (\ref{SGA}) with $\epsilon = 0.005$ to optimize $\delta$. For all CNN based model we use a batch size of 6 and 30 epochs for training the patch. For Segformer, to fit the model in the memory we do a batch size of 1 and just 15 epochs (as there are more frequent gradient updates). The stabilization of Intersection over Union (IoU) decay curves in Fig. \ref{fig:enter-label2} clearly demonstrates negligible benefit of doing additional epochs of patch training. At the end of training we obtain 7 patches (corresponding to 7 models), each with a resolution of $200\times 200$ (patch size is prefixed to 200). Throughout this study we use MIoU \cite{wang2023revisiting} as a metric to quantify segmentation performance.

\subsection{Segmentation performance decay during patch training}
\label{decay}
Since there are only 6 real-time SS models under consideration (excluding Segformer), we study the performance decay of these models as the adversarial patches are optimized over epochs in Fig. \ref{fig:enter-label2}(a). Depending on the structure of each model, every curve presents a unique decay structure. It is interesting to note that MIoU decay curve of even models with same architecture (PIDNet family) can be widely different based on model complexity. Around epoch 15, PIDNet-L shows accelerated decay in performance presenting an interesting behavior to study further. Additionally, BiSeNetV2 shows a rapid decay in performance demonstrating its susceptibility to adversarial attacks of this nature. In Fig \ref{fig:enter-label2}(b,c,d), we present the class wise decay in performance of PIDNet class of models (-L,-M,-S), quantified using class specific Intersection over Union (IoU). A few interesting things to note here: (i) The performance on `sky' pixels practically remains unaffected. This might have to do with \textit{simplicity} of the class. Model -L however does achieve degradation in performance on `sky' class after epoch 17. (ii) The classes `road' or `sidewalk' shows significant degradation for both -L and -S model. However for -M, its not as drastic. This possibly demonstrates the dependence on model structure. (iii) For -L, class `road' even shows a little improvement after epoch 15. However since most other classes show degradation, overall performance goes down (as seen in panel (a)). (iv) Besides `sky', `vegetation' also shows very limited decay in performance for all 3 models, possibly due to its unique spatial structure.

\begin{figure*}
    \centering
    \includegraphics[width=0.81\linewidth]{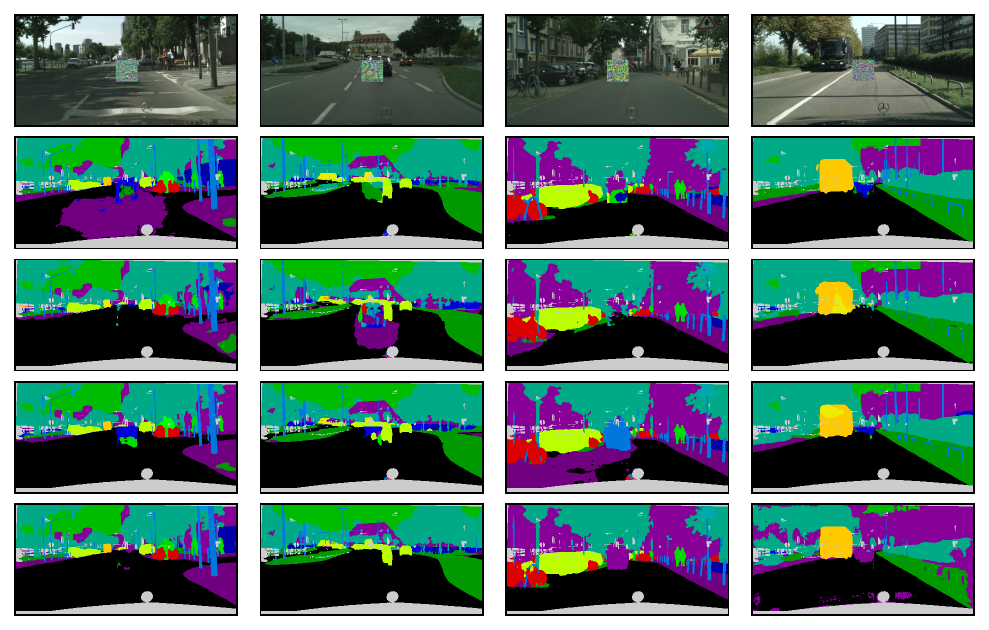}
    \caption{Demonstration of patch performance on 4 random images from Cityscapes-val. Top row shows the images attacked with PIDNet-L, ICNet, BiSeNetV1 and Segformer patches respectively. Rows 2-5 show the predicted segmentation mask from these 4 models (row2:PIDNet-L, row3:ICNet, row4:BiSeNetV1, row5:Segformer) across each of the attacks.}
    \label{fig:enter-label3}
\end{figure*}

\subsection{Patch attack transferability and generalization}
\label{transferability}
Table \ref{tab:cityscape_val} and Fig. \ref{fig:enter-label4} analyze the capability of the adversarial patches to generalize. In Table \ref{tab:cityscape_val}, rows correspond to patch attack trained with respect to a specific model (shown in col 1). Additionally row 0 in-particular shows the case where Cityscapes-val images are attacked by a random noise patch. This random attack provides us the baseline performance for comparing actual adversarial attacks. Columns show the performance of each of these models under each of the attacks. Here we study 3 types of transferability:
\begin{enumerate}
    \item \textit{Transferability across models and architectures (CNNs and ViTs)}: From Table \ref{tab:cityscape_val} we conclude, if a patch is trained with respect to a particular model it is practically only useful for that model. For example, a patch trained on PIDNet-L (row 1), only degrades the performance of PIDNet-L (0.8996 to 0.7914) but the performance of the rest of the models practically remain unaffected (comparing random patch row to PIDNet-L patch row). The same is true for patches trained across all models. For better explanation, in Fig. \ref{fig:enter-label4} we also show the bar plots of drop in MIoU for all 7 models when individual patch attacks are applied to the Cityscapes-val images. Here green bars show positive reduction in MIoU and tiny red bars (visible in some cases) shows negative reduction (increment) in MIoU. Hence for example, if a patch is trained on Segformer (last subplot in Fig. \ref{fig:enter-label4}), only Segformer sees considerable drop in performance and rest of the models are practically unaffected. Additionally, atleast based on the attack discussed, the attacks don't translate well between CNNs and ViTs.
    \item \textit{Transferability across unseen images}: This is demonstrated by training patches on Cityscapes-train and evaluating on Cityscapes-val. Thus from Table \ref{tab:cityscape_val} we can conclude that even on unseen images, the intra-model attacks are effective.
    \item \textit{Transferability across image sizes}: Since the attacks were trained using a crop size of $1024 \times 1024$, but successfully implemented on $1024 \times 2048$ Cityscapes-Val dataset in Table \ref{tab:cityscape_val}, the attacks can be considered effective irrespective of the image size.
\end{enumerate}

To provide the readers with more visual intuition of the effect of adversarial patches, Fig. \ref{fig:enter-label3} shows the performance of 4 models across patches trained with respect to these models. Here the comparison involves CNNs vs CNNs (PIDNet-L vs ICNet vs BiSeNetV1) and CNNs vs ViTs (all CNN models against Segformer). As seen in column 1, with a PIDNet-L patch, only PIDNet-L's prediction is being affected. Again like Fig. \ref{fig:enter-label}, there is a big patch of sidewalk being predicted in the middle of the road. Similarly for ICNet patch (col 2), the effect is most prominent on ICNet prediction (row 3 col 2). The other two models and patches show similar behavior. One very interesting aspect of CNNs vs ViTs we notice here is that CNN based patch attacks mostly influence the prediction in the neighborhood of the patch as seen in subplots (2,1), (3,2) and (4,3). However ViTs due to their token and self-attention structure, propagate the effect to far-off regions in the image. For example in subplot (5,4), we can see the magenta `building' class pixels basically taking over the `sky'/`vegetation' pixels on the left and top.

\section{Discussion and conclusion}
The current research provided a comprehensive overview of the effectiveness of patch based adversarial attacks on real-time semantic segmentation systems from the transferability point of view. For this, we firstly introduced a EOT based formulation of a black-box SS attack that simulates real-time attacks more closely as compared to previous studies \cite{athalye2018synthesizing}. Following this, we considered an experiment suite of 7 different SS models (6 CNNs-based and 1 ViT-based) and studied the Intra-model and Inter-model attack scenarios in a variety of settings. Based on the results obtained, we conclude that patches trained on a particular model don't translate well to other models. In-fact, the model being attacked might stay totally unaffected if the attack was not particularly trained for it. However, due to EOT, the attacks do translate well to unseen images in different cities, under different lighting and angles. Moving forward, our next immediate goal is to make these patches more robust with respect to models, so that they are simultaneously effective across multiple architectures and thus can be used in development of more robust and secure segmentation piplelines in autonomous vehicles.

\bibliographystyle{IEEEtran}
\bibliography{references}

\end{document}